# Normalization of Relative and Incomplete Temporal Expressions in Clinical Narratives

Weiyi Sun[1], Anna Rumshisky[2], Ozlem Uzuner[1],

[1] University at Albany, SUNY, Albany, NY. [2] University of Massachusetts, Lowell, MA.

| Weiyi Sun | Anna Rumshisky | Ozlem Uzuner |
| --- | --- | --- |
| 1400 Washington Ave., | 198 Riverside St., | 1400 Washington Ave., |
| Draper 114B, | Olsen Hall, | Draper 114A, |
| Albany, NY 12222 | Lowell, MA 01854 | Albany, NY 12222 |

Corresponding Author: Weiyi Sun
Email: wsun2@albany.edu
Tel: 860-534-1416

# Normalization of Relative and Incomplete Temporal Expressions in Clinical Narratives

## Abstract

**Objective**  To improve the normalization of relative and incomplete temporal expressions (RI-TIMEXes) in clinical narratives.

**Methods**  We analyze the RI-TIMEXes in temporally annotated corpora and propose two hypotheses regarding the normalization of RI-TIMEXes in the clinical narrative domain: the anchor point hypothesis and the anchor relation hypothesis. We annotate the RI-TIMEXes in three corpora to study the characteristics of RI-TMEXes in different domains. This informed the design of our RI-TIMEX normalization system for the clinical domain, which consists of an anchor point classifier, an anchor relation classifier and a rule-based RI-TIMEX text span parser. We experiment with different feature sets and perform error analysis for each system component.

**Results**  The annotation confirmed the hypotheses that we can simplify the RI-TIMEXes normalization task using two multi-label classifiers. Our system achieves anchor point classification, anchor relation classification and rule-based parsing accuracy of 74.68%, 87.71% and 57.2% (82.09% under relaxed matching criteria) respectively on the held-out test set of the 2012 i2b2 temporal relation challenge.

**Discussion**    Experiments with feature sets reveals some interesting findings such as the verbal tense feature does not inform the anchor relation classification in clinical narratives as much as the tokens near the RI-TIMEX. Error analysis shows that underrepresented anchor point and anchor relation classes are difficult to detect.

**Conclusion**    We formulate the RI-TIMEX normalization problem as a pair of multi-label classification problems. Considering only the RI-TIMEX extraction and normalization, the system achieves statistically significant improvement over the RI-TIMEX results of the best systems in the 2012 i2b2 challenge.

## 1. BACKGROUND AND SIGNIFICANCE

Temporal expressions (TIMEXes) are the natural language words or phrases that carry information about time. For example, the phrase "last Friday" in "the patient tripped and fell last Friday" is a TIMEX that indicates when the incident occurred. Narrative texts rely on TIMEXes to present the timeline of a story. In clinical narratives, TIMEXes specify important clinical information such as the duration of symptoms and the frequency of medications. Hence, understanding TIMEXes is a crucial part of any natural language processing (NLP) task that deals with the temporal dimension.

TIMEXes come in various forms. For example, to express the same concept, one can say 'Nov 29, 2013', 'last Friday', 'the 29th', 'the day after Thanksgiving', or 'Jenny's birthday'. To understand such TIMEXes, NLP systems need not only to identify the TIMEXes, but also to normalize all the various forms of expressing 'Nov 29, 2013' into a canonical format.

In many cases, the TIMEX text spans provide sufficient information for normalization (e.g., 'Nov 29, 2013' almost always means the same date in any context). However, this is not the case for two types of TIMEXes: relative and incomplete TIMEXes (collectively referred to as "**RI-TIMEX**" hereafter). **Relative TIMEXes** are phrases whose temporal meanings are stated as relative values against other time points (e.g., "two days before the fall"). **Incomplete TIMEXes** refer to TIMEXes that contain partial information towards their normalized value. For example, the TIMEX in "the lab result at 6am" states the time but we need to refer to the context to determine its calendar date.

RI-TIMEXes are abundant in narrative texts – on average, they constitute 26% of all TIMEXes in clinical narratives, 32% of TIMEXes in historical narratives and 54% of TIMEXes in newswire articles (see Section 3 for details). Normalizing RI-TIMEXes is a challenging task. In the 2012 i2b2 temporal relation challenge,[1] a shared-task challenge on clinical narratives with a TIMEX extraction and normalization track, the top 10 systems achieved an average normalization accuracy of 0.32 in RI-TIMEXes in comparison to the overall TIMEX normalization accuracy of 0.67.

RI-TIMEX normalization requires two pieces of reference information apart from its text span: an anchor point and an anchor relation. The **anchor point** is the TIMEX that the RI-TIMEX refers to. The **anchor relation** is the temporal relation between the RI-TIMEX and its anchor point. It shows how the RI-TIMEX relates to the anchor point in the narrative timeline. In this work, we propose and test the assumption that RI-TIMEX normalization can be treated as a pair of two multi-class classification problems. With assumptions established, we evaluate a novel RI-TIMEX normalization system using a novel approach in the clinical narratives. More specifically, we propose that the anchor point and anchor relation of RI-TIMEXes can be formulated as two multi-label classification problems, which differentiates our method from the state-of-the-art rule-based approaches. The proposed RI-TIMEX normalization system is fully informed by the context surrounding the RI-TIMEXes and thus we expect it to achieve better performance in the clinical narrative domain than existing methods.

## 2. EXISTING WORK

### 2.1 TIMEX standards

Various standards exist in the general domain for representing TIMEXes.[3] The most frequently used standard, TIMEX3, adopted in temporal specification languages such as TimeML(add citation), defines four types of TIMEXes: date, time, duration and set (frequency). It also normalizes the TIMEX's value to ISO8601 format. The TIMEX3 standard uses temporal function to indicate RI-TIMEXes and marks the TIMEX IDs of their anchor points. For example, "two weeks from June 7, 2003" contains two TIMEXes: the duration "two weeks" and the date "June 7, 2003"; the temporal function here is that the duration "two weeks" has the start point of the date "June 7, 2003".

In the clinical domain, independent from the TimeML scheme, there are several temporal representation schemes tailored for clinical narratives, including studies based on temporal constraint model [4, 5], and OWL-based time ontology [6]. We refer readers to Sun et al.'s survey paper for more details [3]. The i2b2 temporal challenge developed a temporally-annotated clinical corpus using the TIMEX3 standard.[2] To simplify annotation, i2b2 removed the temporal function attribute in TIMEX3 and, instead, used temporal relations (TLINKs) to indicate the anchoring points of relative TIMEXes. Another major change in the i2b2 version of TIMEX3 annotation is that instead of using a single document creation time (DCT) for each document, i2b2 assigns a section creation date (SECTIME) for each section. In the i2b2 corpus, each discharge summary contains a clinical history section and a hospital course section. The SECTIMEs for clinical history sections are the admission dates, and the SECTIMEs for hospital course sections are the discharge dates.

## 2.2 TIMEX-annotated datasets

Temporally-annotated corpora are available in newswire, historical narrative and clinical domains. This article uses one widely-used representative corpus from each of these domains. In the newswire domain, we study the widely adopted TimeBank data [14] consisting of 183 newswire articles, 63K tokens, and 1,423 TIMEXes. In the historical narrative domain, we study the WikiWars corpus consisting of 22 Wikipedia articles, 119K tokens, and 2,681 TIMEXes.[7] In the clinical domain, we study the i2b2 temporal challenge dataset consisting of 310 discharge summaries, 178K tokens, and 3,844 TIMEXes.[16]

## 2.3 TIMEX learning

Existing works show that TIMEX, especially RI-TIMEX, normalization is a challenging task. In both the TempEval challenges [8, 9, 10] and i2b2 temporal reasoning challenge,[1] the best performing systems achieved F1-measure in the 90s on TIMEX text span identification (TIMEX extraction), approximating or exceeding the inter-annotator agreements in these corpora. In contrast, in the task of TIMEX normalization in the same two challenges, the accuracy of the best performing systems remained in the mid 80s and lower 70s, much lower than the inter-annotator agreement on the same task in these corpora, respectively. The participants of the challenges recognize TIMEX normalization as a difficult task [11]. In particular, the authors of the best performing TempEval system, HeidelTime, concluded in their error analysis for TempEval 3, that "wrongly detected reference times or relations" were one of "the main sources for incorrect value normalization of underspecified expressions".[12]

In the newswire domain, the common strategy to resolve RI-TIMEXes is to anchor every RI-TIMEX to the document creation time (DCT) and use verbal tense and/or lemma indicators (e.g. 'ago', 'prior') to determine the anchor relation.[13, 14, 11, 12, 15] Few temporal information extraction works exist on general corpora other than newswire. Strotgen et al.[16] analyzed the difference between RI-TIMEX anchor points in four domains: newswire, historical narratives, colloquial (a corpus of text messages), and scientific text. Their strategy is to anchor all RI-TIMEXes to DCT in the news, colloquial, and scientific corpora. In historical narratives, they anchor all RI-TIMEXes to previous TIMEXes. A **previous TIMEX** is a TIMEX that appears immediately before the RI-TIMEX in the narrative text. For instance, in the sentence "*on December 7, 1941, the Japanese attacked Pearl Harbor, and the following day the United States declared war on Japan*", the previous TIMEX of the RI-TIMEX, "*the following day*" is "*December 7, 1941*". To determine anchor relations, the authors used tense as an indicator in news, colloquial, and scientific text, (e.g., to assign the relation 'before' for RI-TIMEXes in past-tense sentences, and the relation 'after' for present-tense and future-tense sentences), and always assign 'after' anchor relations for RI-TIMEXes in historical narratives. This strategy achieved value f-measures of 73.8 for the newswire domain, 74.5 for the historical narrative domain, 91.9 for the colloquial domain and 64.7 for the scientific domain. We show in Section 3 that these strategies are not suitable for the clinical domain.

In the clinical domain, the participants in the i2b2 challenge mainly fall into two categories in their RI-TIMEX handling strategy. In the first category, the systems assign SECTIMEs as anchor points to RI-TIMEXes.[17, 18] For instance, one would anchor an RI-TIMEX to the admission date if words such as "admission" and "operation" appear near the RI-TIMEX. In the second category, the participants define a list of keywords, such as "operation" and "date of birth". When any of the keywords appears in the text, the systems associate it with a nearby TIMEX. Finally, for an RI-TIMEX containing certain signal phrases (e.g. "post-operative day #6" and "day of life #5"), the system uses the TIMEX associated with the relevant keyword as the anchor point of the RI-TIMEX.[19, 20, 21, 22] These approaches seem intuitive, but they cannot handle relative TIMEXes that do not contain any signal phrases (e.g. "two days later") or incomplete TIMEXes. These approaches are difficult to generalize. Indeed, the challenge result shows that RI-TIMEX normalization is an unsolved problem in the clinical domain.[15]

## 3. RI-TIMEX ANNOTATION

### 3.1 Assumptions

Finding the anchor points and determining the anchor relations of the RI-TIMEXes are critical steps towards RI-TIMEXes normalization. To understand and compare the characteristics of the anchor points and anchor relations of RI-TIMEXes in different narrative domains, we annotated the RI-TIMEXes in three widely-used temporally-annotated corpora: TimeBank, WikiWars and the i2b2 temporal corpus. We designed our annotation guidelines based on the following assumptions for RI-TIMEX anchor points and anchor relations.

**Anchor point assumptions:** As discussed in Section 2, the existing methods for finding anchor points are limited to 1) assuming a default anchor point; or 2) looking for RI-TIMEXes that contain certain signal phrases. The first method builds on the assumption that when writers narrate, they tend to follow the timeline and speak in regards to the time of when the writing occurs (i.e., DCT). For narratives spanning a longer time period, writers tend to follow the timeline continuance established by the previous TIMEXes. In rare cases do they jump around the timeline without explicitly stating the new time point because this risks losing the readers. The second method suggests that in clinical narratives, there are cases when the writers would use some significant clinical events as a temporal anchor point and keep referring to it. In the above 'post-operative day #6' example, the anchor point is fixed to the operative day even though other post-operative days may have been mentioned between the text that specifies 'operation day' and the RI-TIMEX 'post-operative day #6'. Inspired by the implicit assumptions underlying the two lines of existing works in RI-TIMEX normalization, we propose the following hypothesis regarding RI-TIMEX anchor points: An RI-TIMEX in narrative text usually anchors to one of the following TIMEXes: document (or section) creation time; previous TIMEX; previous absolute TIMEX. A previous absolute TIMEX refers to the non-RI-TIMEX that appears immediately before the RI-TIMEX. Figure 1 shows a snippet of a de-identified discharge summary that serves as an illustration of our hypothesis. The TIMEXes in the text are shown in italic and underscored. The RI-TIMEX, "the next day", is relative to the TIMEX that appears prior to it, the previous TIMEX, '2017-04-26', and thus the value of RI-TIMEX is 2017-04-27. In this case, since the previous TIMEX also happen to be an absolute TIMEX. the RI-TIMEX "the next day" is anchored to both the previous TIMEX and the previous absolute TIMEX. For the RI-TIMEX, "postoperative day two", its previous TIMEX is "the next day", and it's previous absolute TIMEX is "2017-04-26". The anchor point for it is the previous absolute TIMEX. From this example, we show that our previous absolute TIMEX hypothesis is based on the assumption that certain events are significant in the timeline so that their time stamps are usually explicated stated as absolute TIMEXes, which later RI-TIMEXes often refer back to.

> The patient was admitted to XXX on *2017-04-26* and underwent a coronary artery bypass graft times four with left internal mammary artery to left anterior descending.
>
> The patient was weaned *the next day* [2017-04-27] from mechanical ventilation.
>
> On *postoperative day two* [2017-04-28], the patient's hematocrit was noted to be 23.1 ; he was transfused one unit of packed red blood cells as well as given a dose of Lasix .
>
> The Neo-Synephrine was weaned off by *postoperative day number three* [2017-04-29].

| RI-TIMEX | Value | Anchor Point | Anchor Relation |
|---|---|---|---|
| *the next day* | 2017-04-27 | Previous TIMEX/Previous Absolute TIMEX (2017-04-26) | After |
| *postoperative day two* | 2017-04-28 | Previous Absolute TIMEX (2017-04-26) | After |
| *postoperative day number three* | 2017-04-29 | Previous Absolute TIMEX (2017-04-26) | After |

Figure 1. RI-TIMEX Anchoring Example

**Anchor relation assumptions:** Anchor relation is the temporal relation between the RI-TIMEX and its anchor point. When we model time in a continuous linear fashion, we may view TIMEXes as temporal intervals, a period defined by two instantaneous time points on a timeline,[23] Therefore, there can be 13 possible temporal relations between two TIMEXes intervals. [1] We propose that for the purpose of date and time RI-TIMEX resolution, it is sufficient to treat TIMEXes as time points. Thus, our anchor relation assumption is that the anchor relation between an RI-TIMEX and its anchor point can be one of the following: before, after or equal.

**3.2 Annotation**

We annotated the anchor points and anchor relations of RI-TIMEXes in three domains. Although the focus of this paper is RI-TIMEX normalization in the clinical domain, we also briefly describe our annotation results in the newswire and historical narrative corpus to serve as a comparison to the clinical domain data. The comparison can inform us the common and differentiating characteristics of RI-TIMEXes among these domains.

We extract RI-TIMEXes by filtering known formats of absolute TIMEX (e.g. mm/dd/yy format) from all annotated TIMEXes and manually review whether the remaining TIMEXes are RI-TIMEXes are not. For newswire data, we annotated the TimeBank corpus.[24, 10] Among the 1221 "date" and "time" *type TIMEXes, 54% are RI-TIMEXes.* For historical narrative domain, we looked at the WikiWars corpus.[7] Among the 2387 "date" and "time" type TIMEXes, we found 861 (36%) RI-TIMEXes. For clinical narratives, we examined the i2b2 challenge corpus. The corpus contains 310 discharge summaries, split into a training set (190 documents) and a test set (120 documents). There are 1712 and 1282 "date" and "time" type TIMEXes in the training and test set respectively. We found 624 (36%) RI-TIMEXes in the training set.

In the annotation process, we conducted single pass annotation on the newswire, historical narratives and i2b2 training set. To evaluate annotation quality, 50% of the i2b2 testing set was dual annotated. Our annotation results and inter-annotator agreement is shown in Section 5.1

## 4. METHODS

Our RI-TIMEX normalization strategy is to learn the RI-TIMEX anchor point and anchor relation using multi-label classifiers, and to combine these with the information extracted from the RI-TIMEX text span to form the final value of the RI-TIMEX. For instance, to normalize the RI-TIMEX 'post-operative day # six', we first learned that the anchor point, the date of the operation, is 2006-09-16; the anchor relation is 'after'; and finally the normalizing value is 6. We can then add 6 days to the value of the operative date to obtain the value for the RI-TIMEX, 2007-09-22.

## 4.1 System structure

The structure of our proposed TIMEX extraction and normalization system is as follows:

(Insert Figure 2)

Figure 2. Structure of the TIMEX extraction and normalization system

Since the TIMEX extraction and the normalization of absolute TIMEXes are not the main focus of this paper, we use the existing HeidelTime rule-based TIMEX extraction and normalization system for these sub-tasks.[12] The HeidelTime system was developed for the general domain. The out-of-the-box rules in HeidelTime are not suitable for the clinical domain. Without tuning, the TIMEX extraction F1-measure on i2b2 corpus (10-fold cross validation) is 72 and the value accuracy of absolute TIMEXes is 39%. The value accuracy is the percentage of correctly normalized TIMEXes in all correctly extracted TIMEXes. We adapted the extraction and normalization rules using the training set of the i2b2 corpus. After tuning, the extraction F1-measure (10-fold cross validation) is 92 and the absolute TIMEXes value accuracy is 74%.

After extracting TIMEXes and normalizing the absolute TIMEXes, we process the RI-TIMEXes using the following steps:

- **Anchor point classifier.** We train four binary SVM classifiers to learn each anchor point types: admission date, discharge date, previous TIMEX and previous absolute TIMEX. When none of the classifiers returns positive classifications, we use admission date as the default anchor point. When the classifiers return conflicting classifications, we select the class label based on its prevalence in the training set, that is, in the following order of preference: admission date, previous TIMEX, previous absolute TIMEX, discharge date. We used the LibSVM implementation of SVM for this classifier.[25]

- **Anchor relation classifier.** Similarly, we treat the anchor relation problem as another multi-label classification task with labels "before", "after" and "equal/during" using LibSVM.

- **Value normalization.** The last piece of information required to decipher the value of the RI-TIMEX is the meaning conveyed in the text span of the RI-TIMEX. We composed a set of rules to parse the RI-TIMEX spans. The rules parse the numbers (in both digit forms and word forms), and the expressions of units such as weeks, days and months.

- **Integration.** This step combines the outputs of the above components to generate the final value of the RI-TIMEX. In the case where a RI-TIMEX's anchor point is another RI-TIMEX, the integration step runs recursively to find the final value of the target RI-TIMEX.

## 4.2 Feature sets

We experimented with several features sets in the temporal anchor point and anchor relation classifiers. We used chi-square attribute selection in our experiments. To determine the chi-square threshold value, we performed 10-fold cross validation on the training set using all the features. We chose to use 7.88 as the threshold for anchor point classification and 9.58 for anchor relation classification. The feature sets are:

(A). Bag of words in an N-token window before and after the RI-TIMEX as well as the RI-TIMEX text span itself. Ten-fold cross evaluation results indicate that for this data set, choosing N to be any number from 6 to 10 tokens yields better results. We report the results with N=8. The tokens are case normalized. Both unigrams and bigrams are included.

(B). Same bag of words features as (A), where we normalize all numbers, both in digit formats (e.g. '2', "2$^{nd}$", '#2') and in spelled-out word formats (e.g. 'two', 'second'), to a unified number token. In another words, we do not distinguish any numbers from each other – "2" is considered as the same token as "3".

(C). EVENT features. In i2b2 data, any text span that indicates a clinically significant event was annotated with an EVENT tag. There are 6 types of EVENTs: PROBLEM, TREATMENT, TEST, CLINICAL_DEPT, OCCURRENCE and EVIDENTIAL.[2] Some RI-TIMEXes are closely related to EVENTs (e.g., "the day of admission" where "admission" is an i2b2 EVENT). We thus included features of the EVENTs that exist in the same clause of a sentence as the RI-TIMEX. Both EVENT type and normalized EVENT text span are included.

(D). Previous TIMEX features. The annotation shows that many RI-TIMEXes are anchored to the previous TIMEX or previous absolute TIMEX. It is natural to add them as features to inform the prediction of anchor points and anchor relations. We further tested the following ways of feature representation:

1. TYPE attribute (DATE, TIME, DURATION and FREQUENCY) of the previous TIMEX only
2. The bag of words of the previous TIMEX text span.
3. The bag of words of the previous DATE or TIME type TIMEX text span
4. The bag of words of the previous absolute DATE and TIME type TIMEX text span.
5. A flag indicating whether the previous TIMEX is the section time 'admission date', the section time 'discharge date', or neither.

(E). Tense information of the sentence. Sentence tense may inform anchor relation classification. We use the verbal tense of the main verbs as features in our classifiers.

## 5. RESULTS AND DISCUSSION

### 5.1 Annotation results

Table 1 shows the distribution of anchor points and relations in the three domains. An RI-TIMEX may simultaneously anchor to more than one anchor point when the anchor points happen to have the same value. Hence, the percentages of the anchor point distribution for each domain may not add up to 100%.

We found that in TimeBank, 96% of all RI-TIMEXes anchor to the DCTs. In WikiWars corpus, we observe 93% of the RI-TIMEXes anchoring to previous TIMEXes. In contrast, we find that the RI-TIMEXes in the training set of the 2012 i2b2 corpus shows more variety in terms of anchor point. The distribution indicates that the approaches adopted in the existing works will work well in the newswire and historical narrative domains; that is, by assuming DCT and previously mentioned TIMEXes as default anchor points, the systems can correctly anchor the majority of the RI-TIMEXes.[14] However, such a strategy will not work on clinical narratives data because by assuming majority label (i.e. admission date), it can at most correctly anchor 59% of the RI-TIMEXes( 41% if assuming section time, because some of the admission date anchoring happens in the hospital course section and some of the discharge date anchoring happens in the clinical history section, so assuming section time would achieve worse performance than assuming admission date). Interestingly, in all three datasets, at least 95% of all RI-TIMEXes are anchored to at least one of following time points: document creation time (admission or discharge time), previous TIMEX and previous absolute TIMEX. The RI-TIMEXes that are not anchored to one of these time points are mostly in the following categories: 1) annotation error that assigned a wrong TIMEX value to the RI-TIMEX which cannot be anchored to any anchor points. 2) the anchor point never appeared in the text (e.g. "on Feb-8, 20 days after the accident" where the true anchor point of TIMEX "20 days after" should be be date of the accident which never appeared in the text); 3) other infrequent cases where the RI-TIMEXes is another time point mentioned in the text. This confirms our anchor point hypothesis.

In the anchor relation annotation, we found that all three corpora show variety in the temporal anchor relations. The two narrative domain corpora, WikiWars and i2b2 discharge summaries, share similar distribution of anchor relation types, featuring more "After" and "Equal/During" anchor relations, while the newswire corpus differs by showing more 'Before' and 'Equal/During' relations.

We calculated the inter-annotator agreement between two annotators in 50% of the test set. For anchor point annotation, the two annotators agreed on 94.7% of the RI-TIMEX incidences. For the anchor relation annotation, the annotators agreed on 98.4% of the RI-TIMEXes whose anchor points were agreed on.

|  | Types | TimeBank | Wiki Wars | i2b2(Training) | | i2b2(Test) |
|---|---|---|---|---|---|---|
| Anchor points annotation statistics | DCT | 96% | 1% | Admission | 53% | 59% |
|  |  |  |  | Discharge | 16% | 19% |
|  | Previous TIMEX | 2% | 93% |  | 37% | 34% |
|  | Previous absolute TIMEX | 1% | 2% |  | 35% | 28% |
|  | Not in the above types | 1% | 4% |  | 5% | 2% |
| Anchor relations annotation statistics | Before | 38% | 12% | 11% | | 12% |
|  | After | 6% | 55% | 46% | | 46% |
|  | Equal/During | 55% | 30% | 41% | | 41% |
|  | Not in the above types | 1% | 3% | 2% | | 1% |

Table 1. RI-TIMEXes annotation statistics

## 5.2 Feature selection results

The feature selection experiment results are shown in Table 2. We started by including all the features and report the result in the first row. We report accuracies of the classification by label, which is comparable to the end-to-end evaluation results in Section 5.3 . We choose to the use the feature set

(B)+(D1)+(D2) in the final system, because its improvement over feature sets (A) and (B) in discharge date anchor point classification and anchor relation classifications is statistically significant in randomization test. While the results using (B) + (D3) +(D4), (B) + (D3) +(D4) + (D5), and (B) + (D3) +(D4) + (E) are not statistically different from (B)+(D1)+(D2), we choose the (B)+(D1)+(D2) set due to its simplicity.

We found that feature set (C), EVENT features, does not inform the classification of either the anchor point or the anchor relation. Our error analysis shows that most of the events that are related to the RI-TIMEX are already included in the N-token window of the RI-TIMEX. Events further away from the RI-TIMEX add more noise than discriminating information to the classification. We also found that feature set (E), tense features, does not help improve the performance of anchor relation prediction. This is because the majority of verbal tenses are past tense in the discharge summary with exception of few future tense cases in sentences mentioning follow-up plans. Thus, in most cases, the tense information is only indicative of anchor relations when the anchor point is discharge date, which is infrequent. It turns out that the signal words, such as "next" and "prior", in or near the RI-TIMEXes are usually more informative than the tense information in determining the temporal anchor relation.

| Feature Sets | Anchor Points | | | | Anchor Relation | | |
|---|---|---|---|---|---|---|---|
| | Admission (330) | Discharge (101) | Previous TIMEX (228) | Previous absolute TIMEX (221) | After (287) | Before (69) | Equal / During (256) |
| All features | 79.01 | 91.67 | 67.53 | 75.00 | 92 | 72.9 | 89.3 |
| (A) | 74.52 | 90.71 | 65.54 | 76.44 | 91.6 | 75.7 | 89.7 |
| (B) | 76.76 | 91.19 | 64.74 | 77.88 | 93.4 | 71.4 | 84.6 |
| (B) + (C) | 75.80 | 88.62 | 62.34 | 76.44 | 83.6 | 74.3 | 90.5 |
| **(B) + (D1) + (D2)** | **77.56** | **92.47** | **68.91** | **75.16** | **93.4** | **81.4** | **92.1** |
| (B) + (D3) +(D4) | 78.53 | 91.99 | 68.91 | 75.64 | 91.3 | 77.1 | 90.1 |
| (B) + (D3) +(D4) +(D5) | 78.85 | 91.99 | 68.75 | 76.60 | 90.9 | 77.1 | 90.1 |
| (B) + (D1) + (D2) + (E) | 78.85 | 91.67 | 67.47 | 75.16 | 93.4 | 80.0 | 90.1 |

Table 2. 10-fold cross validation in the training set accuracy using different feature sets. The number of instances for each class is shown in parentheses in the first row.

The results also show that the normalized bag of words feature set (B) informs the classification of anchor points but does not help with the classification of anchor relation. Our error analysis shows that the numbers in RI-TIMEX span are sometimes useful in predicting anchoring relations. For example, in some clinical documents, the providers refer to any time after an event but still within 24 hours as "day #1". If a baby was born on the morning of Feb 8, Feb 8 was considered as 'day of life #1' and Feb 9 was considered as occurring on 'day of life #2'. In this case, even though both TIMEXes anchor to the same anchor point, from the anchor relation point of view, the RI-TIMEX "day of life #1" is considered as having an "Equal/During" anchor relation with the anchor point, the date of the birth, while the RI-TIMEX "day of life #2" is considered as having an "After" anchor relation with the anchor point. Thus, normalizing all the numbers to a unified token adds noise to the anchor relation classifier even though it improves the anchor point classifier. The previous TIMEX feature appears to be helpful in both anchor point and anchor relation detection.

## 5.3 End to end results

We report the end-to-end evaluation results on the held-out test set of the 2012 i2b2 corpus in this section. We report the results on the features set (B) + (D1) + (D2). The first two columns of Table 3 show the accuracy of each step of the system. For each step, the accuracy is computed over the RI-TIMEXes that are correctly extracted or classified in the previous step(s). For instance, the anchor point classifier accuracy, 74.68%, means that among all the RI-TIMEXes whose text spans were correctly extracted by the system, the anchor point classifier correctly identified the anchor point of 74.68% of them. For this research, fine tuning and adapting HeidelTime TIMEX extraction rules for better extraction scores was not the main purpose. We refer readers to the 2012 i2b2 participants' papers for more detailed description of TIMEX extraction based on adapted HeidelTime rules.[26, 17]

|  | Overall Accuracy (Recall*) | By Type | By Type Accuracy |
|---|---|---|---|
| Extraction | 82.12%* |  |  |
|  |  | N/A | N/A |
| Anchor point | 74.68% |  |  |
|  |  | Admission date (286) | 91.56% |
|  |  | Discharge date (95) | 50.67% |
|  |  | Previous Timex (166) | 64.42% |
|  |  | Previous Absolute Timex (136) | 77.78% |
| Anchor relation | 87.71% |  |  |
|  |  | Before (58) | 68.97% |
|  |  | After (221) | 90.06% |
|  |  | Equal/During (197) | 90.29% |
| Normalization | 57.2% | | |

Table 3. Accuracy of RI-TIMEX normalization at each step. The number of cases in each class is shown in parentheses next to its class label. (The * value for TIMEX extraction is a recall measure result)

Table 3 also shows the break-down statistics of the anchor point classifier and anchor relations over correctly extracted RI-TIMEXes. The anchor point classifier has a similar performance on the training and test set for previous TIMEX and previous absolute TIMEX type. However, the performance on admission date and discharge date anchor points in the test data are quite different. Our strategy of preferring admission date label in the cases of missing labels, and discriminating against discharge date labels in cases of conflict labels increased the accuracy of admission date anchor points and brought down the accuracy of discharge date type of anchor points.

Similar to the 10-fold cross validation results, the anchor relation classifier achieves better performance than the anchor point classifier on the test set. Recall that 46% and 41% of all RI-TIMEX examples in the training set are 'after' and 'equal/during' relations, and only 11% are 'before' relations. It is not surprising to see that the 'before' accuracy is much lower than the accuracy of the other two relations. The anchor relation classifier achieves an accuracy of 87.71% over all the RI-TIMEXes whose anchor point we correctly identified.

The normalization step in our system uses a set of rules to parse the TIMEX text span and generates a final ISO8601 standard value based on the anchor point, anchor relation prediction and the parsed results. We notice that the performance of this step is poor: only 57.2% of all the RI-TIMEXes whose text spans, anchor points and relations are correctly identified received a correct TIMEX value. Our error analysis shows that an inconsistency in annotation induced by the ambiguity of human language contributed to the poor performance. More specifically, many RI-TIMEXes in the corpus are of the format No. X day post a certain clinical event (e.g. "Post-operative Day #" 'Day of Life #'). It turns out that sometimes the provider will refer to the day of the clinical event (operation, or labor in the above examples) as post event day #1, while other times they refer to the second day as post event day #1. A few of these ambiguities can be clarified when there is an absolute date assigned to the RI-TIMEX (e.g. "Post-operative day #3, 03-14-1998"), but most of the time, it depends on the annotators' interpretation to determine the date. To gauge the effect of this inconsistency, we relax the evaluation criteria so that it allows a +/- 1 day deviation of the post-event day type RI-TIMEXes. Under this criterion, the normalization step accuracy goes up to 82.09%.

Table 4 shows the F1 measure comparison between our system and the 2012 i2b2 participants results under strict (i2b2 evaluation method) and relaxed evaluation. Note that the i2b2 participating systems were not built to optimize the extraction and normalization of RI-TIMEXes, but RI-TIMEXes constitute more than 1/3 of the i2b2 TIMEX track. Table 4 shows how our system compares to these state-of-art systems. When using the relaxed method, our system has a larger gain than existing methods, which suggests that for these ambiguous RI-TIMEXes, our results are closer to truth because of the anchor point and anchor relation classification. A randomization test shows that the improvement is statistically significant under the relaxed evaluation method with a p-value of 0.0001.

|  |  | Extraction Recall | Value accuracy | Value Recall |
|---|---|---|---|---|
| i2b2 evaluation method | i2b2 Avg | 86.07 | 32.04 | 27.64 |
|  | i2b2 top score | 92.31 | 34.62 | 31.96 |
|  | Our result | 90.17 | 37.47 | 33.79 |
| Relaxed evaluation method | i2b2 Avg | 86.07 | 43.54 | 37.51 |
|  | i2b2 top score | 92.31 | 45.42 | 41.93 |
|  | Our result | 90.17 | 53.77 | **48.48** |

Table 4. i2b2 evaluation and Relaxed evaluation comparison between our method and the i2b2 top performing systems (RI-TIMEX only)

## 6. CONCLUSION

Relative and incomplete temporal expressions bring about significant challenges in temporal reasoning due to their dependence on the context. In this paper, we present a novel approach to normalize the value of RI-TIMEXes based on an observation over several temporally annotated narrative corpora from various domains. We show that the approach provides statistically significant normalization results than the existing methods on the held-out test data of the i2b2 corpus. The limitation of this approach includes the requirement of RI-TIMEX anchor point and anchor relation annotation and that it has not been tested on larger TIMEX annotated clinical corpora other than the i2b2 corpus. Due to its simple structure, this method can be readily extended to other corpora or other domains. Further research on adapting this approach to more corpora can be valuable.